\pdfoutput=1
\documentclass[11pt]{article}

\usepackage[preprint]{acl}

\usepackage{times}
\usepackage{latexsym}

\usepackage[T1]{fontenc}
\usepackage[utf8]{inputenc}

\usepackage{microtype}

\usepackage{inconsolata}

\usepackage{graphicx}
\usepackage{booktabs}
\usepackage{subcaption}
\usepackage{amsfonts}
\usepackage{multirow}
\usepackage{amsmath}
\usepackage[normalem]{ulem}
\useunder{\uline}{\ul}{}
\usepackage{listings}
\usepackage{tcolorbox}
\tcbuselibrary{listings}
\usepackage{tabularx}
\usepackage{makecell}

\title{PEFT-Bench: A Parameter-Efficient Fine-Tuning Methods Benchmark}

\author{Robert Belanec$^{\spadesuit}$$^\dagger$, Branislav Pecher$^\dagger$, Ivan Srba$^\dagger$, Maria Bielikova$^\dagger$ \\
$^{\spadesuit}$ Faculty of Information Technology, Brno University of Technology, Brno, Czechia \\
$^\dagger$ Kempelen Institute of Intelligent Technologies, Bratislava, Slovakia\\
\texttt{\{name.surname\}}@kinit.sk\\ }

\begin{document}
\maketitle
\begin{abstract}
Despite the state-of-the-art performance of Large Language Models (LLMs) achieved on many tasks, their massive scale often leads to high computational and environmental costs, limiting their accessibility. Parameter-Efficient Fine-Tuning (PEFT) methods address this challenge by reducing the number of trainable parameters while maintaining strong downstream performance. Despite the advances in PEFT methods, current evaluations remain limited (in terms of evaluated models and datasets) and difficult to reproduce. To bridge this gap, we introduce \textit{PEFT-Bench}, a unified end-to-end benchmark for evaluating diverse PEFT methods on autoregressive LLMs. We demonstrate its usage across \textit{27 NLP datasets} and \textit{7 PEFT methods}. To account for different PEFT training and inference factors, we also introduce the \textit{PEFT Soft Cost Penalties (PSCP)} metric, which takes trainable parameters, inference speed, and training memory usage into account.
\end{abstract}

\section{Introduction}
Large Language Models (LLMs) \citep{radford2019language, achiam2023gpt, dubey2024llama, jiang2023mistral, team2024gemini} achieve remarkable results in many Natural Language Understanding (NLU) and Natural Language Generation (NLG) tasks. This significant increase in performance was possible due to the introduction of the attention mechanism forming the transformer architecture \citep{vaswani2017attention}. To achieve such performance, many currently best-performing models contain a vast number of attention blocks, increasing the number of trainable parameters to millions \citep{devlin-etal-2019-bert}, billions \citep{radford2019language}, or even trillions \citep{fedus2022switch}. With the increase in parameters, the computational and data requirements for training also increased, making the development of LLMs unavailable for many academic institutions or practitioners, as well as producing a bigger carbon footprint. To address this issue, a new paradigm was introduced, denoted as “pre-train, fine-tune”, in which the language model is at first pre-trained on large amounts of raw data and then, in a fully supervised way, adapted to the downstream task \citep{10.1145/3560815}. % This allows research institutes with sufficient resources to train foundational LLMs on enormous datasets and release pre-trained weights. Researchers with limited resources can reuse these shared weights and fine-tune the model on downstream tasks. 
This standard fine-tuning still requires substantial amounts of resources as well as significant quantities of data and storage to store fine-tuned weights for each task.

Parameter-Efficient Fine-Tuning (PEFT) methods aim to tackle these problems by reducing the number of trainable parameters while maintaining the performance on a set of downstream tasks. Over the last years, we have witnessed a steady increase in new PEFT methods being introduced in research works \citep{prottasha2025peft}. However, in practice, due to their efficiency and popularity, predominant LoRA-based methods are commonly employed with autoregressive LLMs, and other PEFT methods are rarely chosen. Moreover, a unified and fully open-sourced evaluation of PEFT methods of different types is still missing. We can summarize the cause within the following points: 1) When using the existing PEFT methods as baselines, the lack of fully functional open-source implementations and essential details on the experimental setup \citep{asai-etal-2022-attempt, shi2024dept, tang2025adept} often prevents fellow researchers from replicating their results. 2) Evaluating only non-autoregressive models often restricts the generalizability of the experiments and makes the method look obsolete in the current era of autoregressive LLMs. 3) Finally, the existing PEFT methods are often evaluated primarily on NLU tasks (especially GLUE and SuperGLUE tasks).

To ultimately improve and unify the evaluation of PEFT methods, we propose a new benchmark \textbf{PEFT-Bench}\footnote{To promote the replicability and usage of our benchmark, we provide a dedicated GitHub repository: \url{https://github.com/kinit-sk/PEFT-Bench}.}, which bridges the gap in the comparison of PEFT methods. We simultaneously introduce the \textbf{PEFT-Factory framework}\footnote{PEFT-Factory source code is available at: \url{https://github.com/kinit-sk/PEFT-Factory}.} ~\cite{belanec2025peft}, which provides a necessary underlying technological support for execution of the PEFT-Bench benchmark.

Our main contributions are as follows:
\begin{itemize}
    \item We introduce the \textit{PEFT-Bench}, an end-to-end benchmark that defines the datasets, metrics, and methodology for evaluating PEFT methods in NLP in a fair and consistent environment. To the best of our knowledge, we are the first to provide such a benchmark in NLU and NLG tasks for PEFT methods with autoregressive LLMs. To demonstrate its use, we evaluate \textit{7 different PEFT methods} in terms of efficiency and stability with limited data on \textit{27 NLP datasets}.
    \item To unify and automate the evaluation, we simultaneously introduce the \textit{PEFT-Factory} framework. The framework allows researchers to extend PEFT-Bench with new PEFT methods, maintaining the same evaluation and comparison setup.
    \item Finally, \textit{we propose the PEFT Soft Cost Penalties (PSCP) metric}, which introduces a number of trainable parameters, memory usage, and inference speed in the final score calculation. PSCP reflects the practical feasibility of deploying PEFT methods in real-world scenarios, where efficiency and resource constraints are often as critical as task performance. 
\end{itemize}

\section{Related Work}
Many new PEFT methods are primarily evaluated on GLUE and SuperGLUE benchmarks and often include the main results only on non-autoregressive models \citep{tang2025adept, zhang2025dynamic, zhang-etal-2025-parameter} (e.g., T5 \citep{raffel2020exploring}, RoBERTa \citep{liu2019roberta}) or include only limited evaluation on autoregressive models \cite{lan-etal-2025-efficient, shi2024dept, belanec2025task}. Although some of the new parametrization-based methods (e.g., SVFT \citep{lingam2024svft}, SVF \citep{sun2025transformersquared}) are already evaluated on autoregressive models and on natural language generation tasks, a more unified and overarching evaluation is still missing. 

While several survey papers on PEFT methods include performance evaluations \citep{ding2023parameter, lialin2023scaling, xu2023parameter}, these are largely limited to encoder-based or encoder–decoder-based models and do not consider autoregressive models (similarly to the evaluation of individual new PEFT methods mentioned in the previous paragraph). A notable exception is the most recent survey \citep{prottasha2025peft}, which does include autoregressive models, but only for a small set of NLG tasks.

Prior benchmarks have primarily focused on evaluating PEFT methods in the computer vision domain \citep{zhang2023zhijian, xin2024v}, leaving benchmarking on textual datasets underexplored. Moreover, such benchmarks often keep their implementation either fully or partially closed-sourced and do not provide a usable interface to rerun them with newly created PEFT methods. 

Additionally, due to low baseline replicability, we have been witnessing a trend in new soft prompt-based methods to draw numerical results directly from the related works (without actually rerunning the experiments) \citep{zhang2025dynamic, tang2025adept}. This practice is often not only methodologically incorrect (if the authors do not perfectly match the original environment), but also weakens the results of the proposed method.

In our work, we aim to mitigate these problems. Our PEFT-Bench is a fully open-source benchmarking approach for PEFT methods with state-of-the-art autoregressive NLP models, and with an easy-to-use interface for fellow researchers to add new methods. In addition, our work is the first systematic end-to-end approach (defining the datasets, metrics, and methodology) to provide a unified evaluation of PEFT methods for NLP.

\section{PEFT-Bench}

\begin{figure*}
  \centering
  \includegraphics[width=1\textwidth]{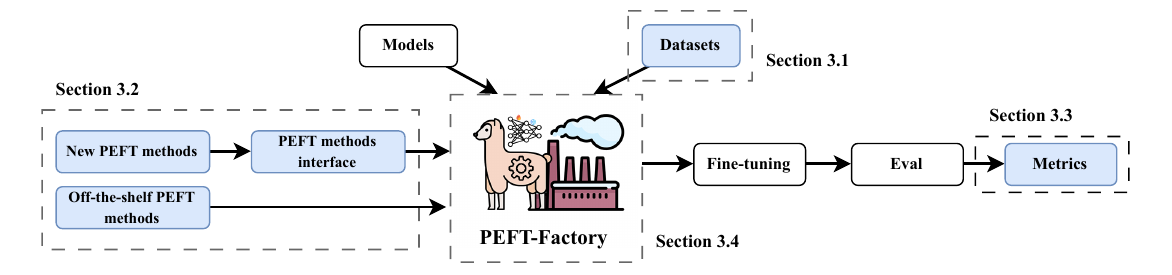}
  \caption{Diagram describing the methodology of \textbf{PEFT-Bench}. Blue components represent our contributions. We design \textbf{PEFT-Factory}, a framework based on LLaMA-Factory \citep{zheng2024llamafactory} backbone to implement off-the-shelf methods from the HuggingFace PEFT library and an easy-to-use interface for new PEFT methods. Using these methods, we train LLaMa on selected datasets, which we have also included in the backbone. After training, we evaluate and compute the metrics for each model, method, and dataset combination.}
  \label{fig:bench}
\end{figure*}

Our PEFT-Bench is constructed in multiple steps, which are visualised in the diagram in Figure \ref{fig:bench}. At a high level, it consists of three main components: 1) a diverse collection of datasets and tasks; 2) representative language models together with a set of PEFT methods; and 3) a set of evaluation metrics.

\subsection{Tasks and Datasets}\label{sec:td}
The PEFT-Bench includes \textbf{27 datasets} representing \textbf{12 unique tasks} in NLU and NLG. These datasets are categorized into 3 main groups: 1) NLU and Reasoning (further subcategorized into GLUE, SuperGLUE, and Others); 2) Math; and 3) Code Generation. An overview of the datasets contained in PEFT-Bench can be seen in Figure \ref{fig:datasets} (detailed information about datasets can be found in the Appendix \ref{app:exp}, Table \ref{app:tab:datasets}). We chose these datasets based on the following criteria: 1) The datasets represent standard tasks, commonly used for fine-tuning in the NLP domain. 2) The datasets provide high task and domain diversity. 3) The size and sequence lengths of a dataset are not overly large (training requires reasonable computational resources).

\begin{figure}
  \centering
  \includegraphics[width=\columnwidth]{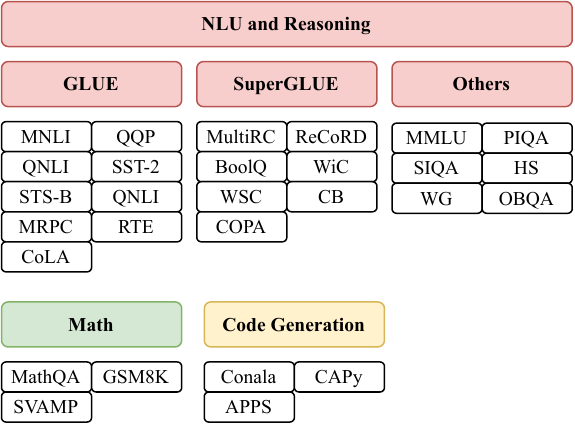}
  \caption{A diagram showing the overview and categorizations of datasets used in PEFT-Bench, totaling 27 datasets categorized into 3 main groups -- NLU and Reasoning, Math, and Code Generation.}
  \label{fig:datasets}
\end{figure}

Current works on new PEFT methods \citep{asai-etal-2022-attempt, shi2024dept, tang2025adept} often evaluate their performance on GLUE \citep{wang2018glue} and SuperGLUE \citep{wang2019superglue} benchmarks. However, since GLUE and SuperGLUE became standard for the evaluation of NLU, they are often used in LLM pre-training and aligning many state-of-the-art models. Due to this, we are also including tasks that are newer and generally harder for LLMs, like code generation or math problem solving. Since we are using supervised fine-tuning to fine-tune an instruction-fine-tuned model, we also include an instruction in each sample (our defined templates of these instructions for each dataset can be found in Appendix \ref{app:exp:templates}).

\subsubsection{Natural Language Understanding and Reasoning Tasks}
Natural language understanding (NLU) is an area of tasks that aim to measure how language models perform in common linguistic problems (e.g., sentiment classification or coreference resolution). Evaluating such tasks became standard when designing a new method or training a new model. This area of tasks is represented by popular benchmarks GLUE and SuperGLUE, which we decided to include in PEFT-Bench.

Reasoning tasks can be either formed as reasoning understanding or as reasoning generation. In our task selection, we include both such forms of reasoning (i.e., reasoning classification and mathematical reasoning generation).

\paragraph{GLUE Benchmark.} PEFT-Bench includes 8 datasets from GLUE, split into 5 tasks, namely \textbf{natural language inference (NLI)} -- \textit{MNLI}~\citep{williams-etal-2018-broad}, \textit{QNLI}~\citep{rajpurkar2016squad}, \textit{RTE}~\citep{dagan2006pascal,bar2006second,giampiccolo2007third,bentivogli2009fifth}; \textbf{paraphrase classification} -- \textit{QQP}~\footnote{\href{https://quoradata.quora.com/First-Quora-Dataset-Release-Question-Pairs}{https://quoradata.quora.com/First-Quora-Dataset-Release-Question-Pairs}}, \textit{MRPC}~\citep{dolan2005automatically}; \textbf{sentiment classification} -- \textit{SST-2}~\citep{socher2013recursive}; \textbf{sentence similarity} -- \textit{STS-B}~\citep{cer-etal-2017-semeval} and \textbf{acceptability classification} -- \textit{CoLA}~\citep{warstadt-etal-2019-neural}.

\paragraph{SuperGLUE Benchmark.} Additionally, PEFT-Bench also includes 7 datasets from SuperGLUE, split into 4 tasks, namely \textbf{natural language inference (NLI)} -- \textit{CB}~\citep{de2019commitmentbank}; \textbf{question answering} -- \textit{MultiRC}~\citep{khashabi2018looking}, \textit{ReCoRD}~\citep{zhang2018record}, \textit{BoolQ}~\citep{clark-etal-2019-boolq}, \textit{COPA}~\citep{roemmele2011choice}; \textbf{word sense disambiguation} -- \textit{WiC}~\citep{pilehvar-camacho-collados-2019-wic} and \textbf{coreference resolution} -- \textit{WSC}~\citep{levesque2011winograd};

All of the GLUE and SuperGLUE datasets are designed as classification problems with the exception of STS-B (which is a regression problem) and ReCoRD (which includes only open questions). To be able to train an autoregressive model that generates text on such classification tasks, we replace numerical class representations with textual labels (e.g., in the QQP dataset, we replace 0 with not\_duplicate and 1 with duplicate).

\paragraph{Other datasets.} PEFT-Bench also includes datasets that are, to the best of our knowledge, not part of any well-known benchmarks or are benchmarks on its own, but are still popular and commonly used to evaluate NLU and reasoning in LLMs, totaling 6 datasets split into 3 tasks, namely \textbf{question answering} -- \textit{MMLU}~\citep{hendrycks2021measuring}, \textit{PIQA}~\citep{bisk2020piqa}, \textit{SIQA}~\citep{sap-etal-2019-social}, \textit{OBQA}~\citep{khot-etal-2019-whats}; \textbf{natural language inference (NLI)} -- \textit{HellaSwag}~\citep{zellers-etal-2019-hellaswag}; \textbf{commonsense reasoning} -- \textit{WinoGrande}~\citep{sakaguchi2021winogrande};

\subsubsection{Math Tasks}
Another popular task that is used for LLM evaluation is mathematical problem-solving. The difficulty of datasets for such evaluation can range from simple question answering to full mathematical reasoning and problem-solving. PEFT-Bench includes 3 datasets for mathematical problem solving split into 3 tasks, namely \textbf{question answering} -- \textit{MathQA}~\citep{amini-etal-2019-mathqa}; \textbf{math word problems} -- \textit{GSM8K}~\citep{cobbe2021training} and \textbf{simple math problems} -- \textit{SVAMP}~\citep{patel-etal-2021-nlp}.

\subsubsection{Code Generation Tasks}
Finally, code generation is another popular task that also contributes to the domain diversity of PEFT-Bench, since it contains mostly code generation questions, which are out-of-domain for the previously mentioned tasks. We mostly focus our selection on the Python language. PEFT-Bench includes 3 datasets for code generation, namely \textit{Conala}~\citep{yin2018learning}, \textit{CodeAlpacaPy}~\citep{codealpaca}, and \textit{APPS}~\citep{hendrycks2021measuring}. 

\subsection{PEFT Methods and Pretrained Models}
\begin{figure}
  \centering
  \includegraphics[width=0.9\columnwidth]{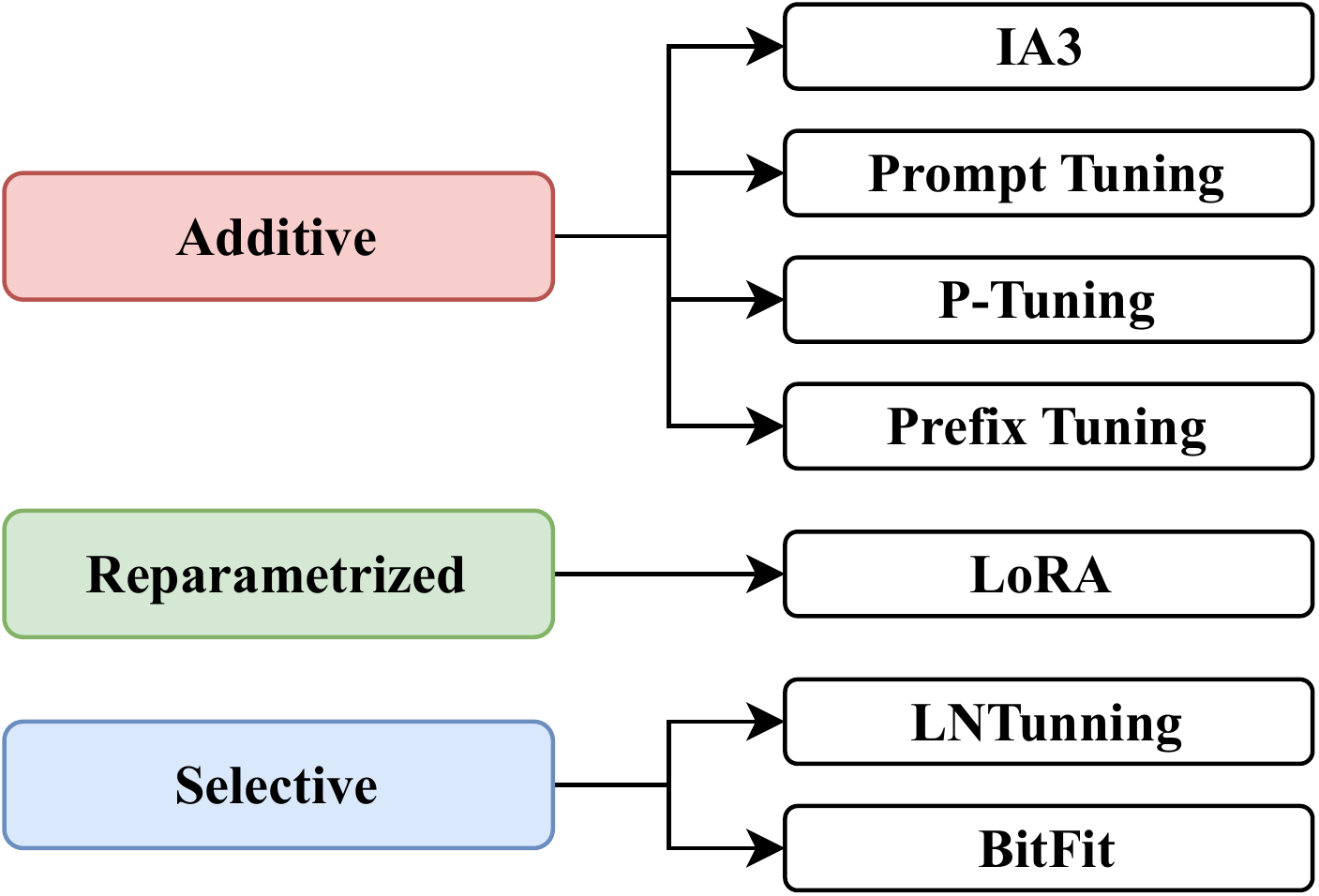}
  \caption{We evaluate methods from additive, reparametrized, and selective PEFT categories. The diagram shows the categorization of each method.}
  \label{fig:methods}
\end{figure}

With the popularization of PEFT, many new methods are being introduced within a short period of time \citep{xu2023parameter, prottasha2025peft}. Therefore, it is computationally expensive to evaluate them all. We design our PEFT method selection in PEFT-Bench to cover \textit{additive PEFT}, \textit{reparametrized PEFT}, and \textit{selective PEFT} categories (the categorization of methods that we have selected can be found in Figure \ref{fig:methods}). To account for each category, we select \textbf{IA$^3$}~\citep{liu2022few}, \textbf{Prompt Tuning}~\citep{lester-etal-2021-power}, \textbf{P-Tuning}~\citep{liu-etal-2022-p}, and \textbf{Prefix Tuning}~\citep{li-liang-2021-prefix} from additive PEFT methods, where IA$^3$ is an adapter-based method, and the rest are soft prompt-based methods. Consequently, we select \textbf{LoRA}~\citep{hu2022lora} from reparametrized PEFT methods and \textbf{LNTuning}~\citep{zhao2024tuning} and \textbf{BitFit}~\cite{ben-zaken-etal-2022-bitfit} from selective PEFT methods. These methods are widely adopted, frequently used as baselines, and together represent one of the most influential and diverse approaches within each PEFT category. We provide a more detailed description of each selected PEFT method in Appendix \ref{app:peft-desc}.

Since the efficiency of PEFT strongly depends on the base model architecture, we select a popular and open-source representative foundation model, \textbf{LLaMa-3-8B-Instruct} \citep{dubey2024llama}, to evaluate all the mentioned PEFT methods. This model also represents a compromise in terms of the number of parameters and performance (we want to train a large enough model to fit within our computational resources).

\subsection{Evaluation metrics}\label{sec:metrics}
To measure task \textbf{performance} in PEFT-Bench, for each dataset, we use standard metrics, namely, F1 for classification tasks and multi-choice QA, accuracy for open answer QA and math problem-solving, and CodeBLEU \citep{ren2020codebleu} for code generation problems. CodeBLEU is an alternative to the standard BLEU metric that also evaluates syntax for different programming languages. 

Current works on PEFT methods often measure solely the accuracy performance or the \textbf{number of trainable parameters}, which makes it harder to compare. Moreover, related works only rarely report the \textbf{memory and inference efficiency} in their results. These are the key factors that may affect the preference for using a specific method over another within a certain computational budget. At the same time, for efficient comparison of benchmarked methods, we need to aggregate performance and such additional PEFT training and inference factors into a single metric. To this end, we propose a new PEFT-bencharking metric called \textbf{PEFT Soft Cost Penalties (PSCP)}. 

We propose to penalize the performance when the costs behind a certain factor are too high. Therefore, we need to scale the performance on the specific task $P_t$ with a penalty for that factor, similarly to the Cobb-Douglass production function \citep{cobb1928theory} for $n$ costs:
\begin{equation}\label{eq:cobb}
    y = P_t \times \prod^n_{i=1} cost_i^{\beta_i},
\end{equation}
where $\beta_i$ is the importance of $cost_i$, and $cost_i^{\beta_i}$ is the final penalty.

In our case, we consider the following factors: 1) the \textbf{number of trainable parameters} $M_p \in \mathbb{N}$; 2) \textbf{inference speed} $M_f \in \mathbb{R}^+$ (i.e., floating point operations); and 3) average \textbf{memory usage during training} $M_m \in \mathbb{R}^+$. Since, these factors are not normalized and cannot be used as penalties on their own, we normalize $M_p$, $M_f$, $M_m$ using reference constants $\frac{M_p}{C_p}$, $\frac{M_f}{C_f}$, $\frac{M_m}{C_m}$, where $C_p \in \mathbb{N}$,  $C_f \in \mathbb{R}^+$ and $C_m \in \mathbb{R}^+$. The constants can be set either to a large value that we would not like to achieve (e.g., number of parameters close to full model training, double the inference floating point operations, or maximum memory available for training), or alternatively, they can also be set to the median or geometric mean of all available $M$ values from a specific factor. If we plug this normalization into Equation \ref{eq:cobb}, we would end up with the intermediate equation \ref{eq:penalty}.

\begin{equation}\label{eq:penalty}
    y = P_t\times(\frac{M_p}{C_p})^{\beta_p}\times(\frac{M_f}{C_f})^{\beta_f}\times(\frac{M_m}{C_m})^{\beta_m} 
\end{equation}

This, however, would rescale the performance $P_t$ to infinity if $M >> C$, therefore we need to bound it to the interval $[0,1)$ with the reciprocal function (i.e., $x^{-1}$). For this reason, we also need to add 1, so that the result of the reciprocal does not go to infinity when $M << C$. The final form of our PSCP metric is given by Equation \ref{eq:pscp}.

\begin{equation}\label{eq:pscp}
    \resizebox{\columnwidth}{!}{%
    $PSCP = P_t\times(1+\frac{M_p}{C_p})^{-\beta_p}\times(1+\frac{M_f}{C_f})^{-\beta_f}\times(1+\frac{M_m}{C_m})^{-\beta_m}$
    }
\end{equation}

The proposed metric is easily extensible if additional factors are of interest. In our case, we consider factors that we want to minimize (i.e., the number of parameters, inference speed, and training memory). If we want to include different factors that we want to maximize (i.e., higher is better), we need to switch the numerator and denominator of the fraction for that cost calculation. 

As the PSCP allows setting the importance of each cost by setting the $\beta$ value, it makes the PSCP metric further configurable and adaptable for different use cases. If we set the importance of the cost $\beta$ to 0, the resulting penalty will be 1, leaving the final PSCP metric unaffected by that factor. If we set the importance of the cost $\beta$ to a number greater than 0, the resulting penalty for that factor will increase (i.e., $cost_i^{\beta_i}$ will get closer to 0). 

Subsequently, if the importance of all factors is set equally, only the magnitude of the PSCP score changes, not the order of methods. However, this occurs only when the performance differences between the compared methods are high. This can be seen in Appendix \ref{app:pscp_importance} in Table \ref{app:tab:pscp_magnitude}. If we want to put more emphasis on a particular cost, we need to set its $\beta$ value greater than that of other costs. This specific behaviour can be seen in Appendix \ref{app:pscp_importance} and Table \ref{app:tab:lora_better}, where we decrease the importance of the number of trainable parameters and, therefore, decrease the penalty, which makes LoRA the method with the highest PSCP score. 

\subsection{PEFT-Factory}

To increase the replicability, modularity, and usability of the PEFT-Bench, we simultaneously introduce \textbf{PEFT-Factory}~\citep{belanec2025peft}, a framework for efficient training of autoregressive LLMs, based on one of the popular LLaMA-Factory LLM training tools \citep{zheng2024llamafactory}. PEFT-Factory includes new off-the-shelf PEFT methods from the HuggingFace PEFT library \citep{peft} and also provides an easy-to-use interface to add newly created PEFT methods to our framework. PEFT-Factory also includes all PEFT-Bench datasets, from which some are specifically adapted for autoregressive generation and classification (e.g., GLUE and SuperGLUE). In addition, PEFT-Factory implements the PSCP metric together with standard metrics and provides automated evaluation and results comparison.

Consequently, PEFT-Factory can be viewed as an instrument of PEFT-Bench that automates the training of different methods in a simple and configurable way. In this manner, PEFT-Factory takes just two configuration files (one for training and one for evaluation) as input, and returns fine-tuned adapter weights alongside the performance metrics. This specific property of PEFT-Factory allows an identical experimental environment during multiple runs (when equivalent hardware configurations are utilized). For this reason, we create a specific training and evaluation configuration for each model and method during the PEFT-Bench execution. 

\section{Experimental Setup}

\begin{table*}[ht]
\resizebox{\textwidth}{!}{%
\begin{tabular}{@{}lr|ccccccccccccccc@{}}
\toprule
\multicolumn{1}{c}{\multirow{3}{*}{Method}} & \multicolumn{1}{c|}{\multirow{3}{*}{\begin{tabular}[c]{@{}c@{}}\# Trainable \\ Parameters\end{tabular}}} & \multicolumn{8}{c}{GLUE (F1/Pearson)} & \multicolumn{7}{c}{SuperGLUE (F1)} \\ \cmidrule(l){3-17} 
\multicolumn{1}{c}{} & \multicolumn{1}{c|}{} & \multicolumn{1}{r}{MNLI} & \multicolumn{1}{r}{QQP} & \multicolumn{1}{r}{QNLI} & \multicolumn{1}{r}{SST-2} & \multicolumn{1}{r}{STS-B} & \multicolumn{1}{r}{MRPC} & \multicolumn{1}{r}{RTE} & \multicolumn{1}{r|}{CoLA} & \multicolumn{1}{l}{ReC} & \multicolumn{1}{l}{MRC} & \multicolumn{1}{l}{BQ} & \multicolumn{1}{l}{WiC} & \multicolumn{1}{l}{WSC} & \multicolumn{1}{l}{CB} & \multicolumn{1}{l}{COPA} \\ \cmidrule(l){3-17} 
\multicolumn{1}{c}{} & \multicolumn{1}{c|}{} & \multicolumn{15}{c}{llama-3-8b-instruct} \\ \midrule
\multicolumn{1}{l|}{Base} & N/A & 59.2 & 0.5 & 80.4 & 92.2 & 67.4 & 78.8 & 69.7 & \multicolumn{1}{c|}{78.3} & 49.4 & 68.0 & 80.8 & 66.6 & 43.6 & 47.0 & 73.8 \\
\multicolumn{1}{l|}{IA$^3$} & 196,608 & {\ul 91.1} & 86.3 & 94.4 & \textbf{96.4} & 88.3 & 86.7 & 84.5 & \multicolumn{1}{c|}{88.9} & 86.0 & 87.0 & {\ul 89.7} & 72.1 & 42.0 & {\ul 68.9} & 96.6 \\
\multicolumn{1}{l|}{Prompt Tuning} & 409,600 & 59.9 & 0.7 & 78.6 & 91.3 & 59.3 & 81.4 & 68.7 & \multicolumn{1}{c|}{76.8} & 38.9 & 75.0 & 77.0 & 65.1 & 5.6 & 41.3 & 79.5 \\
\multicolumn{1}{l|}{Prefix Tuning} & \multicolumn{1}{l|}{34,177,536} & 48.0 & 12.7 & 92.1 & 91.6 & 82.8 & 38.8 & 72.5 & \multicolumn{1}{c|}{45.5} & 26.9 & 45.5 & 72.2 & 60.1 & 0.0 & 49.8 & 7.5 \\
\multicolumn{1}{l|}{P-Tuning} & \multicolumn{1}{l|}{53,130,752} & 85.7 & 78.2 & 81.6 & 94.7 & 3.1 & {\ul 89.4} & 0.0 & \multicolumn{1}{c|}{85.8} & 80.4 & 85.3 & 86.4 & 70.8 & 0.0 & 6.0 & 92.3 \\
\multicolumn{1}{l|}{LoRA} & 14,680,064 & {\ul 91.1} & \textbf{88.2} & \textbf{96.1} & 95.9 & \textbf{90.7} & \textbf{91.0} & {\ul 85.7} & \multicolumn{1}{c|}{\textbf{89.7}} & \textbf{90.3} & \textbf{88.8} & \textbf{91.0} & \textbf{75.2} & \textbf{53.6} & \textbf{83.9} & \textbf{97.8} \\
\multicolumn{1}{l|}{LNTuning} & 266,240 & \textbf{91.2} & {\ul 87.1} & 95.1 & {\ul 96.1} & {\ul 89.8} & 86.8 & 84.4 & \multicolumn{1}{c|}{{\ul 89.3}} & {\ul 88.2} & {\ul 87.3} & 89.7 & {\ul 74.2} & {\ul 48.4} & 67.6 & {\ul 97.7} \\
\multicolumn{1}{l|}{BitFit} & 163,840 & 91.0 & 86.0 & {\ul 95.4} & \textbf{96.4} & {\ul 89.8} & 88.6 & \textbf{87.0} & \multicolumn{1}{c|}{88.6} & 88.1 & 87.2 & {\ul 90.5} & 66.5 & 36.1 & 67.7 & 96.6  \\ \bottomrule
\end{tabular}%
}
\caption{Results of PEFT-Bench with different PEFT methods on NLU tasks from GLUE and SuperGLUE benchmarks. We measure Pearson correlation for STS-B and F1 for others. The best scores are in \textbf{bold}, and the second-best scores are {\ul underlined}.}
\label{tab:results-glue-superglue}
\end{table*}

In all datasets, we utilize 10\% randomly selected from the training set as a validation set and the original validation set as a test set for evaluation. During the validation phase, we measure validation loss and save the weights of the best validation loss for every 5\% of the training steps. We train for 10 epochs with a batch size of 4. We utilize the AdamW \citep{loshchilov2018decoupled} optimizer with a 0.1 warmup ratio, a weight decay of 1.0e-5, a learning rate of 5.0e-5, and a cosine learning rate scheduler. If not specified, we keep the training hyperparameters equal for all of the experiments and change only the PEFT method hyperparameters.

For IA$^3$, we apply adapters for key, value, and downsample feedforward modules of the transformer architecture. For Prompt Tuning, we fine-tune the soft prompt with a length of 100 tokens and initialize it randomly from the vocabulary of the trained model. For P-Tuning, we also fine-tune the soft prompt with a length of 100 tokens, but we reparametrize the soft prompts with an LSTM model with a hidden size of 768. For Prefix Tuning, we fine-tune the soft prompt with a length of 32 tokens, and we project the soft prompt with a small MLP with a hidden size of 512. For LoRA hyperparameters, we set the rank to 16, alpha to 16, and dropout to 0.05. Similarly to IA$^3$, we also reparametrize the key, value, and downsample matrices of the transformer architecture. For LNTuning, we leave all of the settings to their default values (the modules are selected based on the model architecture). For BitFit, we add the bias terms to the query and value matrices. For the stability experiments, we rerun training for 5 different random seeds, and we train each method for 20 epochs on each dataset. 

As the \textit{base model} for all PEFT methods, we adopt LLaMa3-8B-Instruct (containing exactly 8,030,261,248 trainable parameters) since this model is widely adopted, open-source, and supported by a strong community ecosystem. In addition, LLaMa3-8B-Instruct  is a representative of modern instruction-tuned LLMs and provides a fair basis for benchmarking.

To calculate the PSCP metric, we calculate $P_t$ as an average of the performance metric (depending on the particular dataset) across all 5 runs. Since all performance metrics for each dataset are in the range $<0,1>$, we can average the individual performance scores into a single value $P_t$ in a fair manner (as is already done in established benchmarks such as GLUE and SuperGLUE). We set $\beta$ for all factors to 1. For the efficiency factors, we set $M_p$ as the number of trainable parameters of the PEFT model (counting the parameters that are enabled in gradient descent). Consequently, $M_f$ is set as the difference between base model FLOPs and PEFT model FLOPs. Lastly, $M_m$ is set as an average maximum memory across all runs.

We set all reference constants $C$ to maximum values that would still be meaningful for training PEFT methods. More specifically, we set the number of trainable parameters $C_p = 5\times10^8$, as some of the smallest LLMs (e.g., Qwen2.5-0.5B~\citep{Yang2024Qwen2TR}) contain 0.5 billion parameters, and if PEFT methods were to train such a number of parameters, it may be more beneficial to perform a full fine-tuning instead. We set the number of floating-point operations during inference $C_f = 10$ TFLOPs, approximately twice the number of the base model floating-point operations, as going beyond such a number will make inference with PEFT methods unacceptable. Finally, we set the training memory $C_m = 94$ GB, as it is typically the maximum memory available on high-end GPU cards (e.g., Nvidia H100 NVL GPU). More details about the implementation and individual PEFT methods are provided in Appendix \ref{app:exp}.

\section{PEFT-Bench Results}\label{sec:results}

In PEFT-Bench, we first evaluate the PEFT methods on popular NLU datasets from GLUE and SuperGLUE, which can be found in Table \ref{tab:results-glue-superglue}. Consequently, we evaluate the PEFT methods on the other, math, and code generation tasks, with the results being shown in Table \ref{tab:results-others}. 

\begin{table*}[ht]
\centering
\resizebox{\textwidth}{!}{%
\begin{tabular}{@{}lr|cccccccccccc@{}}
\toprule
\multicolumn{1}{c}{\multirow{3}{*}{Method}} & \multicolumn{1}{c|}{\multirow{3}{*}{\begin{tabular}[c]{@{}c@{}}\# Trainable \\ Parameters\end{tabular}}} & \multicolumn{6}{c|}{Others} & \multicolumn{3}{c|}{Math} & \multicolumn{3}{c}{Code Generation} \\ \cmidrule(l){3-14} 
\multicolumn{1}{c}{} & \multicolumn{1}{c|}{} & MMLU & PIQA & SIQA & HS & WG & \multicolumn{1}{c|}{OBQA} & MQA & GSM8K & \multicolumn{1}{c|}{SVAMP} & Conala & CAPy & APPS \\ \cmidrule(l){3-14} 
\multicolumn{1}{c}{} & \multicolumn{1}{c|}{} & \multicolumn{12}{c}{llama-3-8b-instruct} \\ \midrule
\multicolumn{1}{l|}{Base} & N/A & 57.1 & 46.0 & 70.3 & 62.2 & 9.6 & \multicolumn{1}{c|}{78.3} & 26.8 & \textbf{79.2} & \multicolumn{1}{c|}{56.7} & \textbf{31.2} & 32.7 & \textbf{21.6} \\
\multicolumn{1}{l|}{IA$^3$} & 196,608 & 64.6 & 86.0 & 79.2 & 91.7 & 81.4 & \multicolumn{1}{c|}{83.7} & 44.3 & 68.3 & \multicolumn{1}{c|}{85.3} & 23.9 & 32.0 & 18.6 \\
\multicolumn{1}{l|}{Prompt Tuning} & 409,600 & 49.0 & 57.4 & 43.2 & 23.6 & 0.0 & \multicolumn{1}{c|}{65.8} & 4.7 & 73.6 & \multicolumn{1}{c|}{55.7} & {\ul 30.1} & 32.1 & {\ul 20.9} \\
\multicolumn{1}{l|}{Prefix Tuning} & 34,177,536 & 15.4 & 45.3 & 32.0 & 24.9 & 27.5 & \multicolumn{1}{c|}{50.1} & 15.5 & 37.8 & \multicolumn{1}{c|}{85.7} & 27.5 & 9.1 & 15.2 \\
\multicolumn{1}{l|}{P-Tuning} & \multicolumn{1}{l|}{53,130,752} & 60.5 & 80.3 & 69.3 & 65.9 & 0.0 & \multicolumn{1}{c|}{79.7} & 21.4 & 49.2 & \multicolumn{1}{c|}{30.7} & 29.9 & 32.3 & 18.6 \\
\multicolumn{1}{l|}{LoRA} & 14,680,064 & 63.1 & \textbf{88.9} & \textbf{81.5} & \textbf{94.8} & \textbf{88.2} & \multicolumn{1}{c|}{\textbf{87.7}} & \textbf{49.4} & {\ul 69.1} & \multicolumn{1}{c|}{\textbf{88.0}} & 28.9 & \textbf{35.0} & 20.7 \\
\multicolumn{1}{l|}{LNTuning} & 266,240 & \textbf{65.8} & 86.4 & 81.0 & 92.1 & {\ul 81.9} & \multicolumn{1}{c|}{{\ul 85.3}} & {\ul 44.6} & 68.3 & \multicolumn{1}{c|}{85.7} & 26.4 & 32.6 & 18.6 \\
\multicolumn{1}{l|}{BitFit} & 163,840 & {\ul 64.7} &  {\ul 88.4} & {\ul 81.4} & {\ul 93.6} & 79.1 & \multicolumn{1}{c|}{84.1} & 43.8 & 68.3 & \multicolumn{1}{c|}{{\ul 86.7}} & 25.9 & {\ul 33.0}  & 17.9 \\ \bottomrule
\end{tabular}%
}
\caption{Results of PEFT-Bench with different PEFT methods on QA, mathematical problem solving, and code generation tasks. We use accuracy for math problems, CodeBLEU \citep{ren2020codebleu} for code generation problems, and F1 for the others.}
\label{tab:results-others}
\end{table*}

To properly account for different efficiency aspects of PEFT methods, we also evaluate them with the PSCP metric (introduced in Section \ref{sec:metrics}). We gather costs for the number of parameters, inference FLOPs, and training memory, and apply the function from Equation \ref{eq:pscp}. We report the PSCP results in Table \ref{tab:results-pscp} together with the cost values.

\begin{table}[ht]
\centering
\resizebox{\columnwidth}{!}{%
\begin{tabular}{@{}l|lllll@{}}
\toprule
PEFT Method & $P_{avg}$ & $cost_p^{-1}$ & $cost_f^{-1}$ & $cost_m^{-1}$ & PSCP \\ \midrule
IA$^3$ & 74.7 & 1.0 & 0.97 & 0.77 & 55.62 \\
Prompt Tuning & 50.0 & 1.0 & 0.86 & 0.69 & 29.49 \\
Prefix Tuning & 45.9 & 0.94 & 1.0 & 0.77 & 33.18 \\
P-Tuning & 51.7 & 0.9 & 0.86 & 0.73 & 29.26 \\
LoRA & 80.1 & 0.97 & 1.0 & 0.77 & 60.08 \\
LNTuning & 77.8 & 1.0 & 1.0 & 0.77 & \textbf{60.19} \\
BitFit & 75.3 & 1.0 & 1.0 & 0.80 & {\ul 60.10} \\
\bottomrule
\end{tabular}%
}
\caption{Results of the PEFT-Bench across different PEFT methods. LNTuning achieves a slightly better PSCP score than LoRA as it receives a lower penalty for the number of parameters.}
\label{tab:results-pscp}
\end{table}

\begin{figure*}[t]
    \centering
    \includegraphics[width=\textwidth]{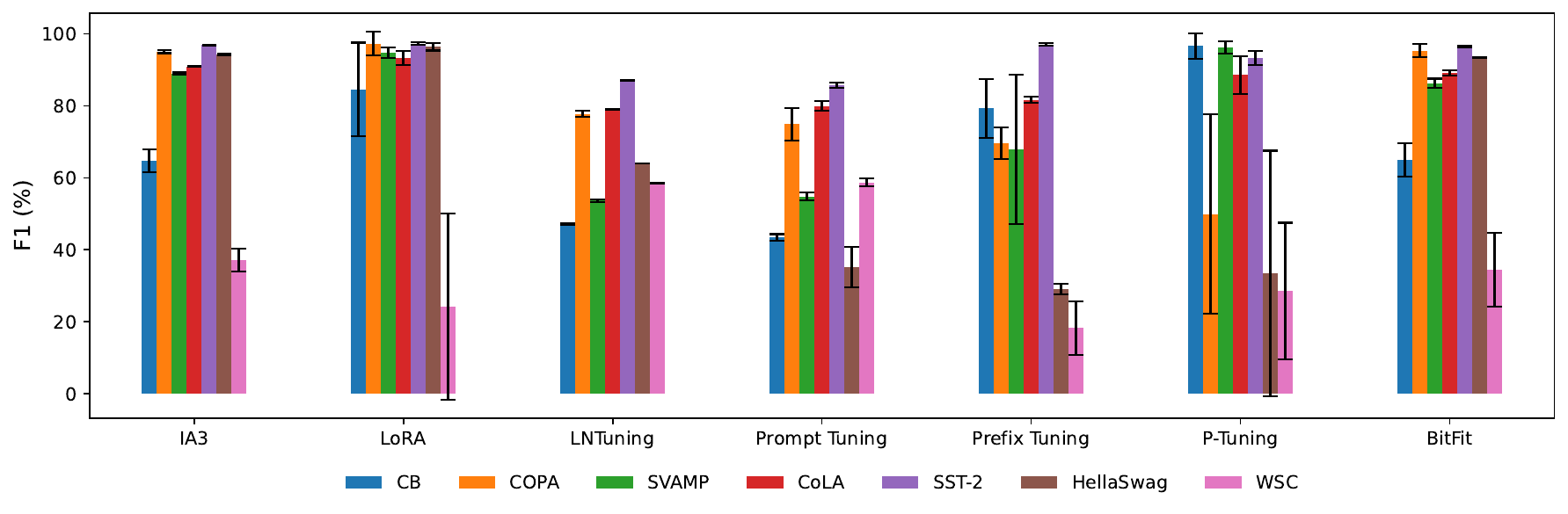}
    \caption{Bar chart showing the stability of different PEFT methods on 4 low-resource datasets. IA$^3$ achieves the lowest standard deviation on average across all datasets, and LoRA is less stable than other methods with CB datasets, while also achieving a better average score. Additionally, P-Tuning generally achieves the worst results in terms of performance and stability. For numerical results, see Table \ref{tab:app:results-stability} in Appendix \ref{app:additional_results}.}
    \label{fig:results-stability}
\end{figure*}

\paragraph{LoRA achieves better performance, but BitFit and LNTuning are more efficient.}
From the performance evaluation results in Tables \ref{tab:results-glue-superglue} and \ref{tab:results-others}, we can see that LoRA achieves the highest performance on many tasks. However, we can also see that a more parameter-efficient methods, BitFit and LNTuning, that train a smaller number of parameters and spend less GPU memory during training, were able to achieve better or on par performance in MNLI, SST-2, RTE and MMLU datasets. This efficiency difference was captured by our PSCP metric in Table \ref{tab:results-pscp}, as it balanced the final result and offset the BitFit and LNTuning methods due to the lower number of parameters and lower memory consumption during training (compared to LoRA).

\paragraph{PEFT learn the task structure but damages the correctness in math problem solving and code generation.} The results in Table \ref{tab:results-others} show a common pattern that PEFTs achieve better performance than the base model in classification or QA tasks. However, for problems that require more generation and reasoning, like GSM8K, Conala, CodeAlpacaPy, or APPS, we are witnessing that the PEFT actually harms the performance. This problem is, however, not affected by the math domain, since the performance on the MathQA and SVAMP datasets is generally increased after parameter-efficient fine-tuning. This may imply that other methods of alignment, like instruction tuning or reinforcement learning, are better suited for problems with longer sequences that require reasoning or chain-of-thought. Additionally, the trainable parameters may still be relatively low, for the model to enhance their reasoning capabilities during supervised fine-tuning.

\paragraph{Soft prompt-based methods are harder to train.} The results in Tables \ref{tab:results-glue-superglue} and \ref{tab:results-others} show that the soft prompt-based methods (i.e., Prompt Tuning, Prefix Tuning, and P-Tuning) perform generally worse compared to other methods, and in some cases, they achieve an F1 score of 0, mostly because they collapse into predicting only a single class. To achieve better results, more fine-grained per-dataset hyperparameter fine-tuning would probably be required. This suggests that out-of-the-box usability and versatility across different datasets are also important factors in designing an efficient method.

Additionally, we have observed memory spikes during training of the Prompt Tuning method. These spikes typically occurred during the first 1,000 steps of training, and they were significantly less noticeable when training on smaller amounts of data. After that, the memory usage was stable for the rest of the training. We believe this may be caused by the direct addition of a soft prompt into the input of an autoregressive model.

\paragraph{Stability experiments.} Finally, we also evaluate the stability of different PEFT methods. Since stability is not directly associated with the efficiency of PEFT methods, we decided to evaluate it separately from the previously-described performance and efficiency results. The results of the stability experiments can be seen in Figure \ref{fig:results-stability}. We evaluated the stability of individual PEFT methods on both, low (CB, COPA, SVAMP, CoLA, SST-2, HellaSwag, and WSC) as well as high resource (SST-2) datasets. Based on the results, the method that achieved the lowest standard deviations on average was IA$^3$. Moreover, we can see that most of the evaluated PEFT methods achieve relatively stable results on small data, except for LoRA on the CB and WSC dataset, where the standard deviations are larger compared to other methods. 

A separate problem can be seen with P-Tuning, which is highly unstable within our experiment setup. This stability problem makes P-Tuning unusable with this setup and would probably require further hyperparameter tuning.

\section{Conclusion}\label{sec:conclusion}
In our work, we introduce PEFT-Bench, the first unified and comprehensive benchmark for parameter-efficient fine-tuning methods for NLP, and PEFT-Factory, as an underlying technological framework supporting PEFT-Bench. PEFT-Bench includes 27 datasets from various NLP problems and domains, implements support for off-the-shelf PEFT methods, simplifies the evaluation of new custom PEFT methods, and finally, introduces 7 PEFT methods as ready-to-use state-of-the-art baselines. We also propose the PSCP metric to properly incorporate efficiency factors like parameters, inference time, and memory usage into the evaluation.

In our comparison, we show that performance on downstream tasks should not be the only condition when evaluating PEFT methods, and sometimes, more efficient PEFT methods, like BitFit and LNTuning, can compensate for the performance with their efficiency. Additionally, we found that soft prompt-based methods are generally harder to train and require extensive hyperparameter tuning.

We view PEFT-Bench as a long-term, evolving initiative that will be continuously updated with new features and the addition of evaluations for different PEFT methods. The current state of PEFT-Bench, together with up-to-date results, is available at its GitHub repository\footnote{\url{https://github.com/kinit-sk/PEFT-Bench}}. In the future, we plan to use this benchmark to evaluate factors that affect the performance of multi-task PEFT methods.

\section*{Acknowledgments}
This work was partially funded by the European Union NextGenerationEU through the Recovery and Resilience Plan for Slovakia under the projects No. 09I01-03-V04-00006 and 09I01-03-V04-00068; and by the European Union, under the project LorAI - Low Resource Artificial Intelligence, GA No. \href{https://doi.org/10.3030/101136646}{101136646}.

Part of the research results was obtained using the computational resources procured in the national project funded by the Ministry of Education, Youth and Sports of the Czech Republic through the e-INFRA CZ (ID:90254); and the national project National competence centre for high performance computing (project code: 311070AKF2) funded by ERDF, EU Structural Funds Informatization of Society, Operational Program Integrated Infrastructure.

\section*{Limitations}
The evaluation of PEFT methods is primarily done on English datasets, which can potentially limit the PEFT-Bench to the English language. However, we believe that there is no specific benefit of evaluating PEFT methods for multilingual settings, as they are independent of the language of the fine-tuning data (unlike pre-trained LLMs, which were pre-trained on data of a specific language).

We also run PEFT-Bench on a single model -- LLaMa3-8B-Instruct. This may limit the generalizability of PEFT-Bench, but, as with language, PEFT methods are not really dependent on the model architecture, and their performance would generally correlate with the model's performance.

Finally, we evaluate 7 different PEFT methods, with the majority being from the additive PEFT and soft prompt-based category. To keep experiments computationally feasible, we rely on the evaluated PEFT methods with limited hyperparameter tuning. Thanks to PEFT-Factory, the whole benchmark is easily extendable and can add their own PEFT method(s).

\section*{Ethical Considerations}
The experiments in this paper were conducted with publicly available datasets MNLI, QQP, QNLI, SST2, STS-B, MRPC, RTE, CoLA, MultiRC, ReCoRD, BoolQ, WiC, WSC, CB, COPA, MMLU, PIQA, SIQA, HellaSwag, WinoGrande, OpenBookQA, MathQA, GSM8K, SVAMP, Conala, CodeAlpacaPy, and APPS, citing the original authors. As we were unable to determine the licenses for all used datasets, we have opted to use them in the limited form possible, adhering to the terms of use of the GLUE and SuperGLUE benchmarks. As the datasets are commonly used in other related works and have been published in scientific works that underwent an established review process, we do not check for the presence of any offensive content, as it was already removed by the authors of these publicly available datasets. In addition, we do not utilize any personally identifiable information or offensive content, and we do not perform crowdsourcing in any form for data annotation. To our knowledge, we are not aware of any potential ethical harms or negative societal impacts of our work, apart from the ones related to the field of Machine Learning (i.e., the use of computational resources that are consuming energy and producing heat with indirect CO2 emission production). We follow the license terms for the LLaMa-3-8B-Instruct model we use -- all models and datasets allow their use as part of the research. As we transform conditional generation into the classification problem (generating only labels), in most cases, we minimize the problem of generating offensive or biased content.

\paragraph{Impact Statement: CO2 Emissions Related to Experiments.} The experiments in this paper require a significant amount of GPU computing resources as we train and evaluate 1 model over multiple random initializations (5) for different methods (6) and datasets (27). Overall, the experiments, including evaluations (which did not require training, but still used GPU resources for inference) and preliminary experiments (which are not reported in the scope of our work), were conducted using a private infrastructure, which has a carbon efficiency of 0.432 kgCO$_2$eq/kWh. Approximately 3000 hours of computation were performed on hardware of type A100 PCIe 40GB (TDP of 250W). Total emissions are estimated to be 9.24 kgCO$_2$eq, of which 0 percent were directly offset. Whenever possible, we tried to reduce the computational costs.

\bibliography{ref}

\appendix

\section{Experiment Setup: Further Details}
\label{app:exp}
This appendix contains extended details about our experiment setup and training of PEFT methods for PEFT-Bench.

\subsection{Detailed Dataset Information}\label{app:exp:datasets}
Table \ref{app:tab:datasets} shows a detailed comparison of information about datasets used in PEFT-Bench. In some cases, the dataset does not contain classes, which means that it either contains open questions (in case of question answering) or it is a generation task.

\begin{table*}[t]
\centering
\resizebox{\textwidth}{!}{%
\begin{tabular}{l|llrrr}
\toprule
\multicolumn{1}{c|}{Area}                             & \multicolumn{1}{c|}{Dataset}      & \multicolumn{1}{c|}{Task}                         & \multicolumn{1}{c|}{\#Classes} & \multicolumn{1}{c|}{\#Train} & \multicolumn{1}{c}{\#Test} \\ \midrule
\multirow{24}{*}{NLU and Reasoning}                   & \multicolumn{5}{c}{\textit{GLUE Benchmark}}                                                                                                                                        \\ \cmidrule{2-6} 
                                                      & \multicolumn{1}{l|}{MNLI}         & \multicolumn{1}{l|}{Natural language inference}   & \multicolumn{1}{r|}{3}         & \multicolumn{1}{r|}{392,702} & 9,815                      \\
                                                      & \multicolumn{1}{l|}{QQP}          & \multicolumn{1}{l|}{Paraphrase classificaiton}    & \multicolumn{1}{r|}{2}         & \multicolumn{1}{r|}{363,846} & 40,430                     \\
                                                      & \multicolumn{1}{l|}{QNLI}         & \multicolumn{1}{l|}{Natural language inference}   & \multicolumn{1}{r|}{2}         & \multicolumn{1}{r|}{104,743} & 5,463                      \\
                                                      & \multicolumn{1}{l|}{SST-2}        & \multicolumn{1}{l|}{Sentiment classification}     & \multicolumn{1}{r|}{2}         & \multicolumn{1}{r|}{67,349}  & 872                        \\
                                                      & \multicolumn{1}{l|}{STS-B}        & \multicolumn{1}{l|}{Sentence similarity}          & \multicolumn{1}{r|}{N/A}       & \multicolumn{1}{r|}{5,749}   & 1,500                      \\
                                                      & \multicolumn{1}{l|}{MRPC}         & \multicolumn{1}{l|}{Paraphrase classificaiton}    & \multicolumn{1}{r|}{2}         & \multicolumn{1}{r|}{3,668}   & 408                        \\
                                                      & \multicolumn{1}{l|}{RTE}          & \multicolumn{1}{l|}{Natural language inference}   & \multicolumn{1}{r|}{2}         & \multicolumn{1}{r|}{2,490}   & 277                        \\
                                                      & \multicolumn{1}{l|}{CoLA}         & \multicolumn{1}{l|}{Acceptability classification} & \multicolumn{1}{r|}{2}         & \multicolumn{1}{r|}{8,551}   & 1,043                      \\ \cmidrule{2-6} 
                                                      & \multicolumn{5}{c}{\textit{SuperGLUE Benchmark}}                                                                                                                                   \\ \cmidrule{2-6} 
                                                      & \multicolumn{1}{l|}{MultiRC}      & \multicolumn{1}{l|}{Question answering}           & \multicolumn{1}{r|}{2}         & \multicolumn{1}{r|}{27,243}  & 4,848                      \\
                                                      & \multicolumn{1}{l|}{ReCoRD}       & \multicolumn{1}{l|}{Question answering}           & \multicolumn{1}{r|}{N/A}       & \multicolumn{1}{r|}{100,730} & 10,000                     \\
                                                      & \multicolumn{1}{l|}{BoolQ}        & \multicolumn{1}{l|}{Question answering}           & \multicolumn{1}{r|}{2}         & \multicolumn{1}{r|}{9,427}   & 3,270                      \\
                                                      & \multicolumn{1}{l|}{WiC}          & \multicolumn{1}{l|}{Word sense disambiguation}    & \multicolumn{1}{r|}{2}         & \multicolumn{1}{r|}{5,428}   & 638                        \\
                                                      & \multicolumn{1}{l|}{WSC}          & \multicolumn{1}{l|}{Coreference resolution}       & \multicolumn{1}{r|}{2}         & \multicolumn{1}{r|}{554}     & 104                        \\
                                                      & \multicolumn{1}{l|}{CB}           & \multicolumn{1}{l|}{Natural language inference}   & \multicolumn{1}{r|}{3}         & \multicolumn{1}{r|}{250}     & 56                         \\
                                                      & \multicolumn{1}{l|}{COPA}         & \multicolumn{1}{l|}{Question answering}           & \multicolumn{1}{r|}{2}         & \multicolumn{1}{r|}{400}     & 100                        \\ \cmidrule{2-6} 
                                                      & \multicolumn{5}{c}{\textit{Others}}                                                                                                                                                \\ \cmidrule{2-5}
                                                      & \multicolumn{1}{l|}{MMLU}         & \multicolumn{1}{l|}{Question answering}           & \multicolumn{1}{r|}{4}         & \multicolumn{1}{r|}{99,842}  & 1,531                      \\
                                                      & \multicolumn{1}{l|}{PIQA}         & \multicolumn{1}{l|}{Question answering}           & \multicolumn{1}{r|}{2}         & \multicolumn{1}{r|}{16,113}  & 1,838                      \\
                                                      & \multicolumn{1}{l|}{SIQA}         & \multicolumn{1}{l|}{Question answering}           & \multicolumn{1}{r|}{3}         & \multicolumn{1}{r|}{33,410}  & 1,954                      \\
                                                      & \multicolumn{1}{l|}{HellaSwag}    & \multicolumn{1}{l|}{Natural language inference}   & \multicolumn{1}{r|}{4}         & \multicolumn{1}{r|}{39,905}  & 10,042                     \\
                                                      & \multicolumn{1}{l|}{WinoGrande}   & \multicolumn{1}{l|}{Commonsense reasoning}        & \multicolumn{1}{r|}{2}         & \multicolumn{1}{r|}{40,398}  & 1,267                      \\
                                                      & \multicolumn{1}{l|}{OBQA}         & \multicolumn{1}{l|}{Question answering}           & \multicolumn{1}{r|}{4}         & \multicolumn{1}{r|}{4,957}   & 500                        \\ \midrule
\multicolumn{1}{c|}{\multirow{3}{*}{Math}}            & \multicolumn{1}{l|}{MathQA}       & \multicolumn{1}{l|}{Question answering}           & \multicolumn{1}{r|}{5}         & \multicolumn{1}{r|}{29,837}  & 4,475                      \\
\multicolumn{1}{c|}{}                                 & \multicolumn{1}{l|}{GSM8K}        & \multicolumn{1}{l|}{Math word problems}           & \multicolumn{1}{r|}{N/A}       & \multicolumn{1}{r|}{7,473}   & 1,319                      \\
\multicolumn{1}{c|}{}                                 & \multicolumn{1}{l|}{SVAMP}        & \multicolumn{1}{l|}{Simple math problems}         & \multicolumn{1}{r|}{N/A}       & \multicolumn{1}{r|}{700}     & 300                        \\ \midrule
\multicolumn{1}{c|}{\multirow{3}{*}{Code generation}} & \multicolumn{1}{l|}{Conala}       & \multicolumn{1}{l|}{Text-to-code}             & \multicolumn{1}{r|}{N/A}       & \multicolumn{1}{r|}{2,379}   & 500                        \\
\multicolumn{1}{c|}{}                                 & \multicolumn{1}{l|}{CodeAlpacaPy} & \multicolumn{1}{l|}{Text-to-code}             & \multicolumn{1}{r|}{N/A}       & \multicolumn{1}{r|}{8,477}   & 942                        \\
\multicolumn{1}{c|}{}                                 & \multicolumn{1}{l|}{APPS}         & \multicolumn{1}{l|}{Text-to-code}             & \multicolumn{1}{r|}{N/A}       & \multicolumn{1}{r|}{5,000}   & 5,000                      \\ \bottomrule
\end{tabular}%
}
\caption{Comparison of datasets used in PEFT-Bench. Each dataset has its own task assigned, totaling 12 unique tasks. The comparison contains the number of classes (if applicable), train sizes, and test sizes for each dataset.}
\label{app:tab:datasets}
\end{table*}

\subsection{Templates used for datasets}\label{app:exp:templates}
In this section, we provide instruction prompt templates that we have used while training for all of the datasets from GLUE (in Table \ref{app:tab:templates-glue}), SuperGLUE (in Table \ref{app:tab:templates-superglue}), Others (in Table \ref{app:tab:templates-others}), and Math and Code Generation (in Table \ref{app:tab:templates-math-code}).

\begin{table*}[!tbh]
\centering
\small
\resizebox{0.95\textwidth}{!}{%
\begin{tabularx}{\textwidth}{@{}lX@{}}
\toprule
\textbf{Dataset} & \textbf{Template} \\ \midrule
MNLI   & \begin{tabular}[c]{@{}l@{}}Classify the following premise and hypothesis pair into entailment, neutral, and contradiction classes. \\ Respond only with the corresponding class.\\\\ \{premise\} \\\\ \{hypothesis\} \\\\ \{label\} \end{tabular} \\ \midrule

QQP    & \begin{tabular}[c]{@{}l@{}}Classify the following question pair into duplicate and not\_duplicate classes. \\ Respond only with the corresponding class.\\\\ \{question1\} \\\\ \{question2\} \\\\ \{label\} \end{tabular} \\ \midrule

QNLI   & \begin{tabular}[c]{@{}l@{}}Classify the following question and sentence pair into entailment and not\_entailment classes. \\ Respond only with the corresponding class.\\\\ \{question\} \\\\ \{sentence\} \\\\ \{label\} \end{tabular} \\ \midrule

SST-2   & \begin{tabular}[c]{@{}l@{}}Classify the following sentence into negative and positive classes. \\ Respond only with the corresponding class.\\\\ \{sentence\} \\\\ \{label\} \end{tabular} \\ \midrule

STS-B   & \begin{tabular}[c]{@{}l@{}}Assign a similarity score from 0 to 5 to the following sentence pair. \\ Respond only with the corresponding score. \\\\ \{sentence1\} \\\\ \{sentence2\} \\\\ \{label\} \end{tabular} \\ \midrule

MRPC   & \begin{tabular}[c]{@{}l@{}}Classify the following sentence pair into not\_equivalent and equivalent classes. \\ Respond only with the corresponding class.\\\\ \{sentence1\} \\\\ \{sentence2\} \\\\ \{label\} \end{tabular} \\ \midrule

RTE    & \begin{tabular}[c]{@{}l@{}}Classify the following sentence pair into entailment and not\_entailment classes. \\ Respond only with the corresponding class.\\\\ \{sentence1\} \\\\ \{sentence2\} \\\\ \{label\} \end{tabular} \\ \midrule

CoLA   & \begin{tabular}[c]{@{}l@{}}Classify the following sentence into unacceptable and acceptable classes. \\ Respond only with the corresponding class.\\\\ \{sentence\} \\\\ \{label\} \end{tabular} \\
\bottomrule
\end{tabularx}
}
\caption{Instruction prompt templates for each dataset in the GLUE benchmark.}
\label{app:tab:templates-glue}
\end{table*}

\begin{table*}[!tbh]
\centering
\small
\resizebox{0.95\textwidth}{!}{%
\begin{tabularx}{\textwidth}{@{}lX@{}}
\toprule
\textbf{Dataset} & \textbf{Template} \\ \midrule
MultiRC   & \begin{tabular}[c]{@{}l@{}}Based on the following paragraph, question, and answer, determine whether the answer answers the question. \\ Respond only with the corresponding true or false.\\\\ Paragraph: \{paragraph\} \\\\ Question: \{question\} \\\\ Answer: \{answer\} \\\\ \{label\} \end{tabular} \\ \midrule

ReCoRD   & \begin{tabular}[c]{@{}l@{}}Based on the following query, entities, and passage, replace the @placeholder in the query. \\ Respond only with the correct replacement.\\\\ Query: \{query\} \\\\ Entities: \{hypothesis\} \\\\ Passage: \{passage\} \\\\ \{label\} \end{tabular} \\ \midrule

BoolQ   & \begin{tabular}[c]{@{}l@{}}Based on the passage, respond to the following question with true or false. \\ Respond only with the corresponding true or false.\\\\ \{question\} \\\\ \{passage\} \\\\ \{label\} \end{tabular} \\ \midrule

WiC   & \begin{tabular}[c]{@{}l@{}}Based on the context from the following sentence pair, classify the word into true and false classes, \\ based on its meaning in both of the sentences. Respond only with the corresponding true or false.\\\\ Sentence1: \{sentence1\} \\\\ Sentence2: \{sentence2\} \\\\ Word: \{word\} \\\\ \{label\} \end{tabular} \\ \midrule

WSC   & \begin{tabular}[c]{@{}l@{}}Based on the following sentence, determine whether the pronoun marked with * * is referencing \\ the noun marked with \# \#. Respond only with the corresponding true and false.\\\\ \{sentence\} \\\\ \{label\} \end{tabular} \\ \midrule

CB   & \begin{tabular}[c]{@{}l@{}}Classify the following premise and hypothesis pair into entailment, contradiction, and neutral classes. \\ Respond only with the corresponding class.\\\\ \{premise\} \\\\ \{hypothesis\} \\\\ \{label\} \end{tabular} \\ \midrule

COPA   & \begin{tabular}[c]{@{}l@{}}Based on the following premise and choice pair, select the plausible alternative from the choice pair. \\ Respond only with the corresponding choice1 or choice2.\\\\ Premise: \{premise\} \\\\ Choice1: \{choice1\} \\\\ Choice2: \{choice2\} \\\\ \{label\} \end{tabular} \\
\bottomrule
\end{tabularx}
}
\caption{Instruction prompt templates for each dataset in the SuperGLUE benchmark.}
\label{app:tab:templates-superglue}
\end{table*}

\begin{table*}[!tbh]
\centering
\small
\resizebox{0.93\textwidth}{!}{%
\begin{tabularx}{\textwidth}{@{}lX@{}}
\toprule
\textbf{Dataset} & \textbf{Template} \\ \midrule
MMLU   & \begin{tabular}[c]{@{}l@{}}Based on the question and provided choices, select the right answer. \\ Respond only with the corresponding choices A, B, C, and D.\\\\ Question: \{question\} \\\\ Choices:\\A:\{choice A\}\\B:\{choice B\}\\C:\{choice C\}\\D:\{choice D\} \\\\ \{label\} \end{tabular} \\ \midrule

PIQA   & \begin{tabular}[c]{@{}l@{}}Based on the question and provided solutions, select the right solution. \\ Respond only with the corresponding solution1 or solution2.\\\\ Question: \{question\} \\\\ Solution1: \{solution1\} \\\\ Solution2: \{solution2\} \\\\ \{label\} \end{tabular} \\ \midrule

SIQA   & \begin{tabular}[c]{@{}l@{}}Based on the context, question, and provided choices, select the right answer. \\ Respond only with the corresponding choices A, B, and C.\\\\ Context: \{context\} \\\\ Question: \{question\} \\\\ Choices:\\A:\{choice A\}\\B:\{choice B\}\\C:\{choice C\} \\\\ \{label\} \end{tabular} \\ \midrule

HellaSwag   & \begin{tabular}[c]{@{}l@{}}Based on the sentence and provided endings, select the correct ending that finishes the sentence correctly. \\ Respond only with the corresponding ending1, ending2, ending3, and ending4.\\\\ Sentence: \{sentence\} \\\\ Ending1: \{ending1\}\\Ending2: \{ending2\}\\Ending3: \{ending3\}\\Ending4: \{ending4\} \\\\ \{label\} \end{tabular} \\ \midrule

WinoGrande   & \begin{tabular}[c]{@{}l@{}}Based on the sentence and provided options, select the best option that replaces the '\_' character \\ in the sentence. Respond only with the corresponding option1 or option2.\\\\ Sentence: \{sentence\} \\\\ Option1: \{option1\} \\\\ Option2: \{option2\} \\\\ \{label\} \end{tabular} \\ \midrule

OBQA   & \begin{tabular}[c]{@{}l@{}}Based on the question and provided choices, select the right answer. \\ Respond only with the corresponding choices A, B, C, and D.\\\\ Question: \{question\} \\\\ Choices:\\A:\{choice A\}\\B:\{choice B\}\\C:\{choice C\}\\D:\{choice D\} \\\\ \{label\} \end{tabular} \\
\bottomrule
\end{tabularx}
}
\caption{Instruction prompt templates for each dataset in the Others category of datasets.}
\label{app:tab:templates-others}
\end{table*}

\begin{table*}[!tbh]
\centering
\small
\resizebox{0.95\textwidth}{!}{%
\begin{tabularx}{\textwidth}{@{}lX@{}}
\toprule
\textbf{Dataset} & \textbf{Template} \\ \midrule
MathQA   & \begin{tabular}[c]{@{}l@{}}Based on the problem and provided options, select the right answer. \\ Respond only with the corresponding choices a, b, c, d, and e.\\\\ Problem: \{problem\} \\\\ Options: a ) \{option A\} , b ) \{option B\} , c ) \{option C\} , d ) \{choice D\} , e ) \{option E\} \\\\ \{label\} \end{tabular} \\ \midrule

GSM8K   & \begin{tabular}[c]{@{}l@{}}Answer following math question. The answer is always numerical. \\ Mark your final answer with a newline starting with '\#\#\#\#'.\\\\ \{question\} \\\\ \{answer\} \end{tabular} \\ \midrule

SVAMP   & \begin{tabular}[c]{@{}l@{}}Answer following math question. The answer is always numerical. \\ Respond only with a single equation with parentheses (e.g., (1 + 1) = 2).\\\\ Question: \{question\} \\\\ \{answer\} \end{tabular} \\ \midrule

Conala   & \begin{tabular}[c]{@{}l@{}}Based on the following instruction, generate a valid Python code that answers it.\\\\ \{rewritten\_intent\} \\\\ \{snippet\} \end{tabular} \\ \midrule

CodeAlpacaPy   & \begin{tabular}[c]{@{}l@{}}Based on the following instruction, generate a valid Python code that answers it.\\\\ \{prompt\} \\\\ \{response\} \end{tabular} \\ \midrule

APPS   & \begin{tabular}[c]{@{}l@{}}Based on the following instruction, generate a valid Python code that answers it.\\\\ \{prompt\} \\\\ \{response\} \end{tabular} \\
\bottomrule
\end{tabularx}
}
\caption{Instruction prompt templates for each dataset in the Math and Code Generation category of datasets.}
\label{app:tab:templates-math-code}
\end{table*}

\section{PSCP Importance Configurations}\label{app:pscp_importance}
In this section, we provide an extended example of different configurations for the importance of each cost $\beta$. To evaluate the changes in the magnitude of the importance of each cost $\beta$, we provide results with different magnitudes in Table \ref{app:tab:pscp_magnitude}. The order of the methods can change if the performance differences between two methods are similar but differ only in a particular cost. 

Based on the results from Table \ref{app:tab:pscp_importance}, we can see that with the change of importance for each cost, the order of the results can change (e.g., in Table \ref{app:tab:lora_better}, the LoRA method achieves the best PSCP score, since we do not penalize the parameters cost as much as the other efficiency factors). Each sub-table represents a different use case, where we emphasize different efficiency factors.

\begin{table*}[ht]
\centering
\resizebox{\textwidth}{!}{%
\begin{subtable}[t]{1.3\columnwidth}
\centering
\begin{tabular}{@{}l|lllll@{}}
\toprule
PEFT Method & $P_{avg}$ & $cost_p^{-1}$ & $cost_f^{-1}$ & $cost_m^{-1}$ & PSCP \\ \midrule
IA$^3$ & 74.7 & 1.0 & 0.97 & 0.77 & 55.62 \\
Prompt Tuning & 50.0 & 1.0 & 0.86 & 0.69 & 29.49 \\
Prefix Tuning & 45.9 & 0.94 & 1.0 & 0.77 & 33.18 \\
P-Tuning & 51.7 & 0.9 & 0.86 & 0.73 & 29.26 \\
LoRA & 80.1 & 0.97 & 1.0 & 0.77 & 60.08 \\
LNTuning & 77.8 & 1.0 & 1.0 & 0.77 & \textbf{60.19} \\
BitFit & 75.3 & 1.0 & 1.0 & 0.8 & \underline{60.1} \\
\bottomrule
\end{tabular}%
\caption{}
\end{subtable}

\begin{subtable}[t]{1.3\columnwidth}
\centering
\begin{tabular}{@{}l|lllll@{}}
\toprule
PEFT Method & $P_{avg}$ & $cost_p^{-2}$ & $cost_f^{-2}$ & $cost_m^{-2}$ & PSCP \\ \midrule
IA$^3$ & 74.7 & 1.0 & 0.94 & 0.59 & 41.41 \\
Prompt Tuning & 50.0 & 1.0 & 0.73 & 0.47 & 17.39 \\
Prefix Tuning & 45.9 & 0.88 & 1.0 & 0.6 & 23.98 \\
P-Tuning & 51.7 & 0.82 & 0.73 & 0.53 & 16.56 \\
LoRA & 80.1 & 0.94 & 1.0 & 0.6 & 45.06 \\
LNTuning & 77.8 & 1.0 & 1.0 & 0.6 & \underline{46.57} \\
BitFit & 75.3 & 1.0 & 1.0 & 0.64 & \textbf{47.97} \\
\bottomrule
\end{tabular}%
\caption{}
\end{subtable}
}

\leavevmode
\newline

\resizebox{\textwidth}{!}{%
\begin{subtable}[t]{1.3\columnwidth}
\centering
\begin{tabular}{@{}l|lllll@{}}
\toprule
PEFT Method & $P_{avg}$ & $cost_p^{-3}$ & $cost_f^{-3}$ & $cost_m^{-3}$ & PSCP \\ \midrule
IA$^3$ & 74.7 & 1.0 & 0.91 & 0.45 & 30.83 \\
Prompt Tuning & 50.0 & 1.0 & 0.63 & 0.33 & 10.26 \\
Prefix Tuning & 45.9 & 0.82 & 1.0 & 0.46 & 17.33 \\
P-Tuning & 51.7 & 0.74 & 0.63 & 0.39 & 9.37 \\
LoRA & 80.1 & 0.92 & 1.0 & 0.46 & 33.8 \\
LNTuning & 77.8 & 1.0 & 1.0 & 0.46 & \underline{36.03} \\
BitFit & 75.3 & 1.0 & 1.0 & 0.51 & \textbf{38.29} \\
\bottomrule
\end{tabular}%
\caption{}
\end{subtable}

\begin{subtable}[t]{1.3\columnwidth}
\centering
\begin{tabular}{@{}l|lllll@{}}
\toprule
PEFT Method & $P_{avg}$ & $cost_p^{-4}$ & $cost_f^{-4}$ & $cost_m^{-4}$ & PSCP \\ \midrule
IA$^3$ & 74.7 & 1.0 & 0.89 & 0.35 & 22.95 \\
Prompt Tuning & 50.0 & 1.0 & 0.54 & 0.23 & 6.05 \\
Prefix Tuning & 45.9 & 0.77 & 1.0 & 0.36 & 12.53 \\
P-Tuning & 51.7 & 0.67 & 0.54 & 0.29 & 5.31 \\
LoRA & 80.1 & 0.89 & 1.0 & 0.36 & 25.35 \\
LNTuning & 77.8 & 1.0 & 1.0 & 0.36 & \underline{27.88} \\
BitFit & 75.3 & 1.0 & 1.0 & 0.41 & \textbf{30.56} \\
\bottomrule
\end{tabular}%
\caption{}
\end{subtable}
}

\caption{Results with PSCP metric under equal importance of each cost with different magnitudes. Because the performance differences between BitFit and LNTuning are small, the relative importance of each cost changed the ordering of the methods.}
\label{app:tab:pscp_magnitude}
\end{table*}

\begin{table*}[ht]
\centering
\resizebox{\textwidth}{!}{%
\begin{subtable}[t]{1.3\columnwidth}
\centering
\begin{tabular}{@{}l|lllll@{}}
\toprule
PEFT Method & $P_{avg}$ & $cost_p^{-0.5}$ & $cost_f^{-1}$ & $cost_m^{-1}$ & PSCP \\ \midrule
IA$^3$ & 74.7 & 1.0 & 0.97 & 0.77 & 55.63 \\
Prompt Tuning & 50.0 & 1.0 & 0.86 & 0.69 & 29.5 \\
Prefix Tuning & 45.9 & 0.97 & 1.0 & 0.77 & 34.29 \\
P-Tuning & 51.7 & 0.95 & 0.86 & 0.73 & 30.78 \\
LoRA & 80.1 & 0.99 & 1.0 & 0.77 & \textbf{60.95} \\
LNTuning & 77.8 & 1.0 & 1.0 & 0.77 & \underline{60.21} \\
BitFit & 75.3 & 1.0 & 1.0 & 0.8 & 60.11 \\
\bottomrule
\end{tabular}%
% \caption{Results of PEFT-Bench, if we lower the importance $\beta_p$ of the number of parameters.}
\caption{}
\label{app:tab:lora_better}
\end{subtable}

\begin{subtable}[t]{1.3\columnwidth}
\centering
\begin{tabular}{@{}l|lllll@{}}
\toprule
PEFT Method & $P_{avg}$ & $cost_p^{-1}$ & $cost_f^{-0.5}$ & $cost_m^{-1}$ & PSCP \\ \midrule
IA$^3$ & 74.7 & 1.0 & 0.98 & 0.77 & 56.47 \\
Prompt Tuning & 50.0 & 1.0 & 0.93 & 0.69 & 31.85 \\
Prefix Tuning & 45.9 & 0.94 & 1.0 & 0.77 & 33.19 \\
P-Tuning & 51.7 & 0.9 & 0.93 & 0.73 & 31.63 \\
LoRA & 80.1 & 0.97 & 1.0 & 0.77 & 60.08 \\
LNTuning & 77.8 & 1.0 & 1.0 & 0.77 & \textbf{60.19} \\
BitFit & 75.3 & 1.0 & 1.0 & 0.8 & \underline{60.1} \\
\bottomrule
\end{tabular}%
% \caption{Results of PEFT-Bench, if we lower the importance $\beta_f$ of the inference speed.}
\caption{}
\end{subtable}
}

\leavevmode
\newline

\resizebox{\textwidth}{!}{%
\begin{subtable}[t]{1.3\columnwidth}
\centering
\begin{tabular}{@{}l|lllll@{}}
\toprule
PEFT Method & $P_{avg}$ & $cost_p^{-1}$ & $cost_f^{-1}$ & $cost_m^{-0.5}$ & PSCP \\ \midrule
IA$^3$ & 74.7 & 1.0 & 0.97 & 0.88 & 63.47 \\
Prompt Tuning & 50.0 & 1.0 & 0.86 & 0.83 & 35.53 \\
Prefix Tuning & 45.9 & 0.94 & 1.0 & 0.88 & 37.73 \\
P-Tuning & 51.7 & 0.9 & 0.86 & 0.86 & 34.22 \\
LoRA & 80.1 & 0.97 & 1.0 & 0.88 & \underline{68.37} \\
LNTuning & 77.8 & 1.0 & 1.0 & 0.88 & \textbf{68.41} \\
BitFit & 75.3 & 1.0 & 1.0 & 0.89 & 67.26 \\
\bottomrule
\end{tabular}%
% \caption{Results of PEFT-Bench, if we lower the importance $\beta_m$ of the used training memory.}
\caption{}
\end{subtable}

\begin{subtable}[t]{1.3\columnwidth}
\centering
\begin{tabular}{@{}l|lllll@{}}
\toprule
PEFT Method & $P_{avg}$ & $cost_p^{-2}$ & $cost_f^{-1}$ & $cost_m^{-1}$ & PSCP \\ \midrule
IA$^3$ & 74.7 & 1.0 & 0.97 & 0.77 & 55.59 \\
Prompt Tuning & 50.0 & 1.0 & 0.86 & 0.69 & 29.46 \\
Prefix Tuning & 45.9 & 0.88 & 1.0 & 0.77 & 31.05 \\
P-Tuning & 51.7 & 0.82 & 0.86 & 0.73 & 26.45 \\
LoRA & 80.1 & 0.94 & 1.0 & 0.77 & 58.37 \\
LNTuning & 77.8 & 1.0 & 1.0 & 0.77 & \textbf{60.16} \\
BitFit & 75.3 & 1.0 & 1.0 & 0.8 & \underline{60.08} \\
\bottomrule
\end{tabular}%
% \caption{Results of PEFT-Bench, if we increase the importance $\beta_p$ of the number of parameters.}
\caption{}
\end{subtable}
}

\leavevmode
\newline

\resizebox{\textwidth}{!}{%
\begin{subtable}[t]{1.3\columnwidth}
\centering
\begin{tabular}{@{}l|lllll@{}}
\toprule
PEFT Method & $P_{avg}$ & $cost_p^{-1}$ & $cost_f^{-2}$ & $cost_m^{-1}$ & PSCP \\ \midrule
IA$^3$ & 74.7 & 1.0 & 0.94 & 0.77 & 53.94 \\
Prompt Tuning & 50.0 & 1.0 & 0.73 & 0.69 & 25.27 \\
Prefix Tuning & 45.9 & 0.94 & 1.0 & 0.77 & 33.14 \\
P-Tuning & 51.7 & 0.9 & 0.73 & 0.73 & 25.05 \\
LoRA & 80.1 & 0.97 & 1.0 & 0.77 & 60.08 \\
LNTuning & 77.8 & 1.0 & 1.0 & 0.77 & \textbf{60.19} \\
BitFit & 75.3 & 1.0 & 1.0 & 0.8 & \underline{60.1} \\
\bottomrule
\end{tabular}%
% \caption{Results of PEFT-Bench, if we increase the importance $\beta_f$ of the inference speed.}
\caption{}
\end{subtable}

\begin{subtable}[t]{1.3\columnwidth}
\centering
\begin{tabular}{@{}l|lllll@{}}
\toprule
PEFT Method & $P_{avg}$ & $cost_p^{-1}$ & $cost_f^{-1}$ & $cost_m^{-2}$ & PSCP \\ \midrule
IA$^3$ & 74.7 & 1.0 & 0.97 & 0.59 & 42.71 \\
Prompt Tuning & 50.0 & 1.0 & 0.86 & 0.47 & 20.31 \\
Prefix Tuning & 45.9 & 0.94 & 1.0 & 0.6 & 25.64 \\
P-Tuning & 51.7 & 0.9 & 0.86 & 0.53 & 21.4 \\
LoRA & 80.1 & 0.97 & 1.0 & 0.6 & 46.39 \\
LNTuning & 77.8 & 1.0 & 1.0 & 0.6 & \underline{46.6} \\
BitFit & 75.3 & 1.0 & 1.0 & 0.64 & \textbf{47.99} \\
\bottomrule
\end{tabular}%
% \caption{Results of PEFT-Bench, if we increase the importance $\beta_m$ of the used training memory.}
\caption{}
\end{subtable}
}

\caption{Results with PSCP metric with different importance of each cost $\beta$. Notice how the order of the methods changes in Table (a) when we decrease the importance of the cost for the number of trainable parameters $\beta_p$.}
\label{app:tab:pscp_importance}
\end{table*}

\section{Descriptions of PEFT Methods}\label{app:peft-desc}
In this section, we provide a description of each PEFT method that we evaluate in this paper.

\textbf{IA$^3$} (Infused Adapter by Inhibiting and Amplifying Inner Activation)~\citep{liu2022few} introduces trainable scaling vectors that scale the activations of specific components within the transformer architecture. Instead of updating all model parameters, IA$^3$ inserts learnable vectors into the attention ($l_k$ and $l_v$) and feed-forward modules ($l_f$). The task is framed as learning the conditional probability $Pr(Y|X)$, parameterized by the frozen model weights $\theta$. IA$^3$ augments this with additional parameters $\theta_{IA^3}$, corresponding to the scaling vectors. During fine-tuning, only $\theta_{IA^3}$ is updated, while $\theta$ remains fixed, making the approach parameter-efficient. 

\textbf{Prompt Tuning}~\citep{lester-etal-2021-power} optimizes a soft prompt that is prepended to the input embeddings. The task is modeled as learning the conditional probability $Pr(Y|X)$, where $X$ denotes a sequence of input tokens and $Y$ represents the output tokens corresponding to the class label. In standard classification, this probability $Pr_\theta(Y|X)$ is parameterized by the model weights $\theta$. Prompting extends this by inserting an additional sequence of tokens (the prompt) $P$ before $X$, so the model instead maximizes $Pr_\theta(Y|[P;X])$, while keeping $\theta$ fixed. Prompt tuning further introduces parameters $\theta_P$ that define the prompt itself, and only $\theta_P$ is updated during training.

Simultaneous work \textbf{Prefix-Tunning}~\citep{li-liang-2021-prefix} introduces similar trainable soft prompts that are prepended before key and value matrices of all attention layers. The soft prompts are not directly parametrized but rather generated by a multi-layer perceptron module.

\textbf{P-Tuning}~\citep{liu2023gpt} considers inserting soft prompts before the inputs, using templates, which are essentially concatenated soft prompts with pre-trained embedding layers. P-Tuning also employs bi-directional LSTM to initialize the embeddings of the soft prompts.

Reparametrization is the process of transforming the original model parameters to other (reparametrized) parameters. In the context of PEFT, the transformation often lowers the number of trainable parameters. These reparametrized parameters can be transformed back to the original model parameter space, removing any overhead during inference. Building on the Intrinsic SAID~citep{aghajanyan-etal-2021-intrinsic} method, \textbf{LoRA}~\citep{hu2022lora} introduced two separate matrices $A \in \mathbb{R}^{r \times k}$ and $B \in \mathbb{R}^{d \times r}$ that form the resulting $\Delta W \in \mathbb{R}^{d \times k}$ matrix, given the initial (pre-trained) weight matrix $W_0 \in \mathbb{R}^{d \times k}$.

\textbf{LNTuning} (LayerNorm Tuning)~\citep{zhao2024tuning} adapts models by updating only the parameters of the layer normalization layers, while keeping all other weights frozen. As in standard classification, the task is modeled as learning the conditional probability $Pr(Y|X)$, parameterized by the model weights $\theta$. This enables efficient adaptation with minimal trainable parameters. Additionally, LNTuning does not create any overhead during inference, since it does not add any additional parameters to the model.

\textbf{BitFit} (Bias Fine-tuning)~\cite{ben-zaken-etal-2022-bitfit} fine-tunes only the bias terms of weights in selected modules of transformer architecture. However, in modern LLMs, the bias term is often skipped while designing the model. In our implementation, instead of reusing the pre-trained bias term (as in the original implementation), we replace the linear layer with a new one that also contains the bias term and copy the pre-trained weights. 

\section{Additional results}\label{app:additional_results}
In this section, we provide additional results to complement the main results in the body of the paper. Table \ref{tab:app:results-stability} shows the numerical results from Figure \ref{fig:results-stability} in Section \ref{sec:results}.

\begin{table*}
\resizebox{\textwidth}{!}{%
\begin{tabular}{llllllll}
\toprule
 & CB & COPA & SVAMP & CoLA & SST-2 & HellaSwag & WSC \\
\midrule
IA$^3$ & $64.7 \pm 3.3$ & $95.0 \pm 0.4$ & $89.0 \pm 0.3$ & $90.9 \pm 0.2$ & $96.8 \pm 0.1$ & $94.2 \pm 0.3$ & $37.1 \pm 3.2$ \\
LoRA & $84.6 \pm 13.0$ & $97.3 \pm 3.4$ & $94.8 \pm 1.5$ & $93.3 \pm 1.9$ & $97.3 \pm 0.4$ & $96.4 \pm 1.1$ & $24.2 \pm 25.8$ \\
LNTuning & $47.2 \pm 0.2$ & $77.7 \pm 0.8$ & $53.6 \pm 0.4$ & $79.0 \pm 0.1$ & $87.1 \pm 0.0$ & $64.0 \pm 0.1$ & $58.6 \pm 0.2$ \\
Prompt Tuning & $43.4 \pm 0.9$ & $74.9 \pm 4.5$ & $54.8 \pm 1.1$ & $79.9 \pm 1.4$ & $85.7 \pm 0.8$ & $35.2 \pm 5.7$ & $58.7 \pm 1.0$ \\
Prefix Tuning & $79.3 \pm 8.1$ & $69.6 \pm 4.4$ & $67.8 \pm 20.8$ & $81.7 \pm 0.9$ & $97.0 \pm 0.4$ & $29.1 \pm 1.5$ & $18.3 \pm 7.5$ \\
P-Tuning & $96.6 \pm 3.5$ & $49.9 \pm 27.7$ & $96.2 \pm 1.8$ & $88.6 \pm 5.2$ & $93.3 \pm 1.9$ & $33.4 \pm 34.2$ & $28.5 \pm 19.0$ \\
BitFit & $65.0 \pm 4.6$ & $95.4 \pm 1.8$ & $86.2 \pm 1.3$ & $89.1 \pm 0.7$ & $96.6 \pm 0.2$ & $93.4 \pm 0.1$ & $34.5 \pm 10.2$ \\
\bottomrule
\end{tabular}
}
\caption{Results from stability experiments. These are numerical results from the graph in Figure \ref{fig:results-stability}.}
\label{tab:app:results-stability}
\end{table*}

\end{document}